\documentclass[10pt,twocolumn,letterpaper]{article}

\usepackage{cvpr}              

\usepackage{graphicx}
\usepackage{amsmath}
\usepackage{amssymb}
\usepackage{booktabs}

\usepackage[pagebackref,breaklinks,colorlinks]{hyperref}

\usepackage[ruled,vlined]{algorithm2e}
\newtheorem{theorem}{Observation}
\usepackage[capitalize]{cleveref}
\crefname{section}{Sec.}{Secs.}
\Crefname{section}{Section}{Sections}
\Crefname{table}{Table}{Tables}
\crefname{table}{Tab.}{Tabs.}

\def\cvprPaperID{6705} 
\def\confName{CVPR}
\def\confYear{2022}

\begin{document}

\title{Towards Better Plasticity-Stability Trade-off in Incremental Learning: A Simple Linear Connector}

\author{Guoliang Lin\\
Sun Yat-sen University\\
Guangdong, China\\
{\tt\small lingliang@mail2.sysu.edu.cn}
\and
Hanlu Chu\\
South China Normal University\\
Guangdong, China\\
{\tt\small  hlchu@m.scnu.edu.cn}
\and
Hanjiang Lai~\thanks{Hanjiang Lai is the corresponding author.}\\
Sun Yat-sen University\\
Guangdong, China\\
{\tt\small  laihanj3@mail.sysu.edu.cn}
}
\maketitle

\begin{abstract}
Plasticity-stability dilemma is a main problem for incremental learning, where plasticity is referring to the ability to learn new knowledge, and stability retains the knowledge of previous tasks. Many methods tackle this problem by storing previous samples, while in some applications, training data from previous tasks cannot be legally stored. In this work, we propose to employ mode connectivity in loss landscapes to achieve better plasticity-stability trade-off without any previous samples. We give an analysis of why and how to connect two independently optimized optima of networks, null-space projection for previous tasks and simple SGD for the current task, can attain a meaningful balance between preserving already learned knowledge and granting sufficient flexibility for learning a new task. This analysis of mode connectivity also provides us a new perspective and technology to control the trade-off between plasticity and stability. We evaluate the proposed method on several benchmark datasets. The results indicate our simple method can achieve notable improvement, and perform well on both the past and current tasks. On 10-split-CIFAR-100 task, our method achieves 79.79\% accuracy, which is 6.02\% higher. Our method also achieves 6.33\% higher accuracy on TinyImageNet. Code is available at \href{https://github.com/lingl1024/Connector}{https://github.com/lingl1024/Connector}.
\end{abstract}


\section{Introduction}
In recent years, deep neural networks have been reported promising performance on various tasks. In the dynamic world, the deep model also needs to be updated as new data becomes available. Hence, Incremental Learning (IL)~\cite{delange2021a,mundt2020a} has received much attention, which studies the problem of continually learning from sequential tasks. 

In this paper, we consider the data-free incremental learning~\cite{smith2021always}, where the training samples from previous tasks do not exist. Hence, the main criterion of data-free IL~\cite{douillard2020podnet,rebuffi2017icarl} is that no data from the previous tasks is stored when continually refining the model as new data becomes available. It is a direct cause of catastrophic forgetting problem~\cite{li2019learn}, and the plasticity-stability dilemma~\cite{chaudhry2018riemannian,mermillod2013the} is a more general problem: (1) \textit{plasticity}: the deep model should learn the new knowledge of the current task, and (2) \textit{stability}: it should also preserve the knowledge of previous tasks.

Many algorithms have been proposed to strike a balance between plasticity and stability. An intuitive solution is to generate samples from previous tasks, e.g., ILCAN~\cite{xiang2019incremental} generates samples to preserve the old knowledge. The regularization-based methods use the extra regularization term in the loss function to consolidate previous knowledge, such as EWC~\cite{kirkpatrick2017overcoming} using the Fisher information to calculate each parameter's importance. The architectural methods~\cite{li2019learn} learn the dynamic architecture of the deep network, e.g., DER~\cite{yan2021der} freezes the previously learned representation and dynamically expands the network for new task. The algorithm-based methods learn parameter updating rules to preserve the performance of previous tasks. For example, GEM~\cite{lopez-paz2017gradient} constrains new task updates which do not interfere with the previous knowledge. Adam-NSCL~\cite{wang2021training} updates network parameters in the null space of all previous tasks and achieves a promising performance on remembering previous knowledge. Although Adam-NSCL can preserve previous knowledge very well, the strong null-space projection also hurts the performance of the current task.  

On the other side, many researches have focused on the connectivity in neural network loss landscapes~\cite{frankle2020linear,fort2019large}. 
The previous works~\cite{garipov2018loss} found that two minima of independently trained deep networks can be connected in weight space, where the loss along the path remains low. Further, the recent works \cite{frankle2020linear} and \cite{wortsman2021learning}  showed that there exists a linear path of high accuracy to connect two minima when the networks share only a few epochs of the initialized SGD trajectory. Mode Connectivity SGD (MC-SGD)~\cite{mirzadeh2021linear} is designed for incremental learning, which enforces the final weight linearly connected to all tasks' minima. Although MC-SGD achieves excellent performance, it needs to store previous samples, which contradicts with our problem setting. Hence, an interesting yet challenging question arises: how to build a high-accuracy pathway between previous and current models without previous samples? 

In this paper, we provide a new insight to understand, analyze and build a high-accuracy connector without any previous training samples. To understand why we can linearly connect two minima of previous and current tasks, we first give a simple analysis that provides an upper bound of empirical loss for all tasks. Then, according to the upper bound, we view plasticity and stability as two independent optimization problems of deep neural networks. The two networks are trained to minimize the empirical loss and move towards each other. Finally, we propose a simple linear connector to attain a better balance between these two networks in the light of linear connectivity~\cite{frankle2020linear}. Central to our method is that we uncover a simple way to achieve a better plasticity-stability trade-off, i.e., a simple averaging of two carefully designed networks, which leads to higher accuracy neural network. 

\section{Preliminaries and Related Methods} 

\subsection{Incremental Learning Methods}
We review several categories of the existing deep incremental learning methods for plasticity-stability trade-off. 

\textbf{Regularization-based methods}:  This line of approaches introduce an extra regularization term to balance the trade-off. According to where the regularization term was explicitly applied to, these methods can be further divided into \textit{structural} and \textit{functional} regularization methods~\cite{mundt2020a}. Structural  regularization methods constrain changes on model's parameters. For example, EWC~\cite{kirkpatrick2017overcoming}, SI~\cite{zenke2017continual}, MAS~\cite{aljundi2018memory} and UCL~\cite{ahn2019uncertainty} explicitly added a regularization term to networks' parameters. The functional regularization methods, also known as the distillation-based methods, use the distillation loss between predictions from the previous model and the current model as the regularization term. The representative works include LwF~\cite{li2018learning}, EBLL~\cite{rannen2017encoder}, GD-WILD~\cite{lee2019overcoming}, etc. 

\textbf{Rehearsal methods}: This line of works preserve existing information by replaying data from previous tasks. Some algorithms store a subset of previous data, e.g., iCaRL~\cite{rebuffi2017icarl} and GeppNet~\cite{gepperth2016a}. When the storage space is limited, it is important to find a suitable subset of data that can approximate the entire data distribution, e.g., SER~\cite{isele2018selective} focuses on exemplar selection techniques. Another way to solve this limitation is using the generative modelling approaches~\cite{xiang2019incremental} to generate a lot of samples of previous tasks. For example, DGR~\cite{shin2017continual} is a framework with a deep generative model and a task solving model. 

\textbf{Architectural methods}: These methods modify the underlying architecture to alleviate catastrophic forgetting, e.g., HAT~\cite{serr2018overcoming} proposes a task-based binary masks that preserve previous tasks' information. UCB~\cite{ebrahimi2020uncertainty} uses uncertainty to identify what to remember and what to change. The dynamic growth approaches~\cite{yan2021der} are also proposed, e.g., DEN~\cite{yoon2017lifelong} dynamically expands network capacity when arrival of new task. Learn-to-Grow~\cite{li2019learn} proposes modifying the architecture via explicit neural structure learning.

\textbf{Algorithm-based methods}: These methods carefully design network parameter updating rule, which constrain new task updates that do not interfere the previous tasks.  GEM~\cite{lopez-paz2017gradient} and A-GEM~\cite{chaudhry2018efficient} are two representative works. OWM~\cite{zeng2019continual} is orthogonal weight modification method to overcome catastrophic forgetting. Adam-NSCL~\cite{wang2021training}, which uses the null space of all previous data to remember existing knowledge, achieves an impressive performance on IL task.

Here, we give a brief review of Adam-NSCL.  
We have a $W_{old}$ model trained on the previous data $X_{old}$, and $X_{old}$ is not available when training the new task. To overcome this problem, Adam-NSCL stores the uncentered feature covariance $ \mathcal{X}_{old}= \frac{1}{n_{old}} X_{old}^{\top} X_{old}$ for guaranteeing stability, where $n_{old}$ is the number of data points in $X_{old}$. Then, it uses the SVD result of feature covariance $\mathcal{X}_{old}$ to find the null space of $X_{old}$, denoted as $U_{old}$. We have $\mathcal{X}_{old} U_{old} = 0$ in this way. The projection matrix is obtained as $P_{old} = U_{old} U_{old}^{\top}$.

Now when new data $X_{new}$ is available, the $W_{old}$ can be updated to learn the new task as:
\begin{equation}
 W_{t+1} = W_t - \alpha  P_{old} \cdot g_t,
\end{equation}
where $W_0 = W_{old}$ and $g_t$ is the gradient only calculated on new data. With the null-space projection, we can update the model that can remember the knowledge of $W_{old}$.

Adam-NSCL can preserve  very well the previous knowledge, while the the updates of new task are limited because of the strong null-space projection. Our method can be viewed as an extension of Adam-NSCL, which achieves a better balance model for both previous and current tasks. 



 

 
\subsection{Linear Mode Connectivity}
Optimizing a neural network involves finding a minimum in a high-dimensional non-convex objective landscape, where some forms of stochastic gradient descent (SGD) are used as  optimization methods for learning the parameters of deep network. Since the deep neural network are non-convex, there are many local minima. Given a deep network $\mathcal{F}$ with an initial weight $W_0$, the weight is iteratively updated and the learnt weight at epoch $k$ is denoted as $W_k = Train(\mathcal{F},W_0)$. Two copies of the deep networks are trained (e.g., using different data augmentations or projections), producing two optimized weights $W_k^1 = Train_1(\mathcal{F},W_0)$ and $W_k^2 = Train_2(\mathcal{F},W_0)$.

Recently, a lot of work~\cite{wortsman2021learning,choromanska2015the,dinh2017sharp} has been developed to study the neural network optimization landscape. Many intriguing phenomena have been found. For example, one of the interesting observations~\cite{draxler2018essentially,garipov2018loss} is that there exists a connector between two optima. The loss minima are not isolated. 

\begin{theorem}[Connectivity]
\cite{draxler2018essentially,garipov2018loss} There exists a continuous path between minima of neural network architectures, where each point along this path has a low loss.
\end{theorem}

To find the continuous path, e.g., from $W_k^1$ to $W_k^2$, Draxler~\cite{draxler2018essentially} proposed a method based on Nudged Elastic Band (NEB)~\cite{jonsson1998nudged} to find the smooth and low-loss nonlinear path. Further, \cite{frankle2020linear} showed that two minima can be connected by a low-loss linear path in some cases.

\begin{theorem} [Linear Connectivity]
\cite{frankle2020linear,wortsman2021learning} There exists a linear connector from $W_k^1$ to $W_k^2$ when $W_0$ is not randomly initialization but is trained to a certain spawn epoch. 
\label{ob2}
\end{theorem}

This condition is easy to satisfy. When the optimization trajectory of $W_0$ is shared, the two optima can be connected in a linear path. Inspired by this, MC-SGD~\cite{mirzadeh2021linear} enforces the final weight linearly connected to all tasks' minima. However, MC-SGD stores a small set of previous samples to learn the linear connector. It can not be used in our data-free setting. 
In this paper, we don't rely on experience replay to force connectivity. Instead, we move the previous and current models closer to ensure connectivity.

\section{Method}
In this section, we first give the problem formulation of incremental learning problem. Let sequential incremental learning tasks be denoted as $T_1,T_2,\dots,T_t,\dots$, and each task includes a set of disjoint classes. In the $t$-th task, we are only given the $t$-th training dataset $\mathcal{D}_t=\left\{\left(x_i,y_i\right)\right\}_{i=1}^{N_t}$, where $N_t$ is the number of training samples, and the previous model $W_{1:(t-1)}$. We need to update the previous model $W_{1:(t-1)}$ to a new model $W_{1:t}$ such that two inherent properties should be considered: 1) \textit{stability}: the new model should retain the knowledge of previous $t-1$ tasks, and 2)  \textit{plasticity}: the  ability to learn the new knowledge of the $t$-th task.

We first give an analysis of how to find high-accuracy pathway. Then, according to the upper bound and linear connectivity, we design a simple linear connector. 



\begin{figure}
    \centering
    \includegraphics[scale = 0.3]{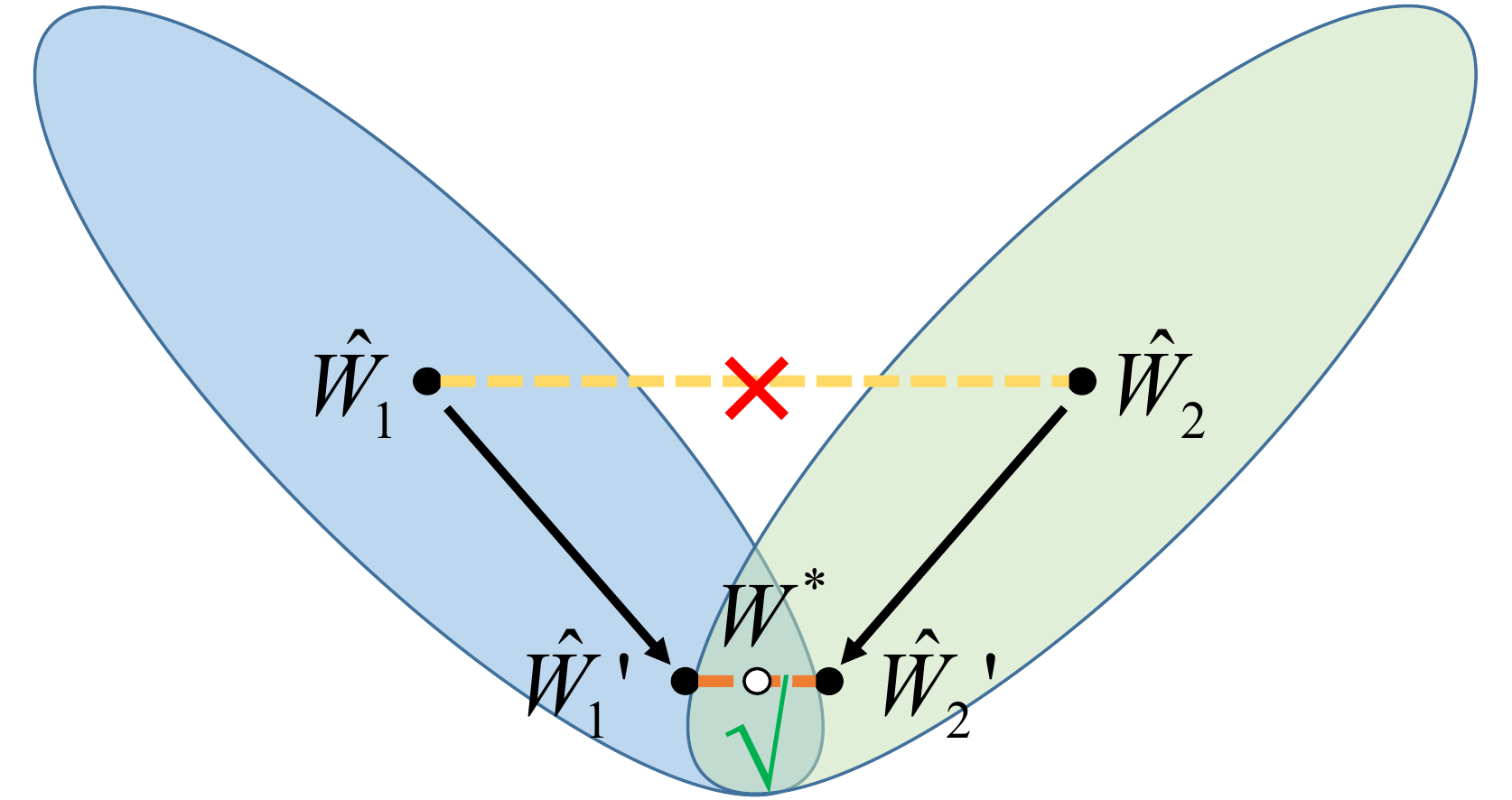}
    \caption{Illustration of how to find high-accuracy pathway, where $\hat W_1$ and $\hat W_2$ are one of optimal weights for task 1 and task 2, respectively. If they are moved closest to each other (e.g., towards the overlapped region), we can find a good linear path.}
    \label{fig:mutual}
\end{figure}


\subsection{How to Build High-accuracy Pathway for All Tasks?}
\label{conlusion}
Suppose that there are $K$ disjoint sequential incremental learning tasks, e.g., two tasks on Figure~\ref{fig:mutual} where the ellipse on the left is the set of optimal weights for task 1 and the right ellipse is the set of optimal models for task 2, can we simply connect any two optimal models of two tasks? For example, the linear connection of $\hat{W}_1$ and $\hat{W}_2$? It may fail since the pathway from $\hat{W}_1$ to $\hat{W}_2$ may result in poor performance. How to find high-accuracy pathway for all tasks is not a trivial problem. Observed on Figure~\ref{fig:mutual}, if we can move $\hat{W}_1$ and $\hat{W}_2$ towards the overlapping region, the high-accuracy path between them will be easier to find. In fact, by moving the optimal weights closest, the upper bound of the empirical loss for all tasks will get smallest. Here comes the theoretical explanation.



Let the optimal or convergent weight for task $i$ be $\hat W_i$, which is only trained on the $i$-th task. The empirical loss for task $i$ is denoted as $L_i(W)$. We aim to find a final weight for all tasks $W^*$ that minimizes the empirical loss for all tasks, e.g., $W^* = arg\min_{W} \sum_{i=1}^{K} L_i (W)$. 

First, we consider one task. For the first task, we can use Taylor expansion to approximate the loss. Following the way in~\cite{mirzadeh2020understanding}, it can be formulated as:
\begin{align}
    & L_1(W^*) \approx L_1(\hat W_1) + (W^* - \hat W_1)^{\top}\nabla L_1(\hat W_1) \notag \\
     & + \frac{1}{2}(W^* - \hat W_1)^{\top}\nabla^2 L_1(\hat W_1)(W^* - \hat W_1) \notag \\
     \le & L_1(\hat W_1) + \frac{1}{2} \lambda_1^{max} \parallel W^* - \hat W_1 \parallel^2,
\end{align}
where $\nabla L_1(\hat W_1) \approx 0$ since $\hat W_1$ is the optimal weight and the gradient's norm vanishes, and $\lambda_1^{max}$ is the maximum eigenvalue of  $\nabla^2 L_1(\hat W_1)$. In the same way, for other tasks, we have:
\begin{gather}
        L_2(W^*) \le L_2(\hat W_2) + \frac{1}{2} \lambda_2^{max} \parallel W^* - \hat W_2 \parallel^2, \notag \\
        \vdots \notag \\
    L_K(W^*) \le  L_K(\hat W_K) + \frac{1}{2} \lambda_K^{max} \parallel W^* - \hat W_K \parallel^2.
\end{gather}
By summing them up, we have:
\begin{equation}
    \sum_{i=1}^{K}L_i(W^*) \le \sum_{i=1}^{K}L_i(\hat W_i) + \frac{1}{2} \lambda^{max} \sum_{i=1}^{K} \parallel W^* - \hat W_i \parallel^2, 
  \label{upperBound1}
\end{equation}
where $\lambda ^{max} = \max(\lambda_1^{max}, \lambda_2^{max},..., \lambda_K^{max})$. Since $\hat W_i$ is the optimal weight for task $i$ ($i=1,\cdots,K$), $L_i(\hat W_i)$ is the minimum. Thus the first term $ \sum_{i=1}^{K}L_i(\hat W_i)$ is also the minimum. Hence, the empirical loss for all tasks, i.e., $\sum_{i=1}^{K}L_i(W^*)$, can be bounded via minimizing the second term $\sum_{i=1}^{K} \parallel W^* - \hat W_i \parallel^2$. It is easy to verify that the optimal weight $W^*_{opt}$ for the second term is: 
\begin{equation}
    W_{opt}^* = \frac{1}{K} \sum\limits_{i=1}^{K} \hat W_i.
\label{centroid}
\end{equation} 
We can see that $W_{opt}^*$ is the centroid or geometric center of $\{\hat W_1, \hat W_2,\cdots, \hat W_K\}$. Putting $W_{opt}^*$ into  $\frac{1}{2} \lambda^{max}  \sum_{i=1}^{K} \parallel W^* - \hat W_i \parallel^2$, we have $\frac{1}{2} \lambda^{max} \sum\limits_{i=1}^{K} \parallel \frac{1}{K}\sum_{j=1}^{K} \hat W_j - \hat W_i \parallel^2 \leq  \lambda^{max} \frac{1}{2K^2} \sum_{i=1}^{K}  \sum_{j=1}^{K} \parallel \hat W_i - \hat W_j \parallel^2$.

Combining the above inequation with Eq.(\ref{upperBound1}), we have 
\begin{align}
\sum_{i=1}^{K}L_i(W_{opt}^*) \le & \sum_{i=1}^{K}L_i(\hat W_i) + \notag \\
&\lambda^{max} \frac{1}{2 K^2} \sum_{i=1}^{K}  \sum_{j=1}^{K} \parallel \hat W_i - \hat W_j \parallel^2.
\label{upperBound2}
\end{align}

The upper bound in Eq.~(\ref{upperBound2}) can give us an interesting perspective of the incremental learning: the empirical loss of all tasks can be bounded by minimizing the sum of empirical loss of each individual task and the sum of squared Euclidean distances between each pair of optimal weights. 

As indicated by the upper bound of Eq.~(\ref{upperBound2}), if we have 1) the previous model $\hat W_{old}$ and current model $\hat W_{new}$ achieve optimal solutions for previous $K-1$ tasks and current $K$-th task, respectively and 2) these two models are moved closest to each other, then we can simply use the linear connection of the two models: $W_{opt}^* = \frac{1}{K} \hat W_{new} +  \frac{1}{K}  \sum_{i=1}^{K-1} \hat W_{old} = \frac{1}{K} \hat W_{new} +  \frac{K-1}{K} \hat W_{old} $ (see Eq.~(\ref{centroid})). So that the upper bound of empirical loss for all tasks would be lowest.

\begin{algorithm}[t]
 \caption{Linear connector for plasticity-stability trade-off}
  \label{alg}
	\KwIn{A set of sequential learning tasks $T_1, T_2, \cdots$, and their training datasets $\mathcal{D}_1,\mathcal{D}_2, \cdots $; A neural network $W$ and learning rate $\alpha$ }
	\BlankLine
	Train the first task to get $W_{1:1} = Train(\mathcal{D}_1)$
	
	\textit{\# compute the null space}
	
    Use the model $W_{1:1}$ and $\mathcal{D}_1$ to obtain feature covariance $\mathcal{X}_{1:1}$ and the null-space projection matrix $P_{1:1}$	 
	
	\For{task $T_t \in \{T_2, T_3, \cdots\}$}{
	\textit{\# init the two networks}
	 
	 Let $\overleftarrow{W}^0_{1:(t-1)} = W_{1:(t-1)}$, $\overrightarrow{W}^0_t = W_{1:(t-1)}$ and $s=0$
	 
	 \While{\textnormal{not converged}}{
         Sample a mini-batch $\{X,Y\}$ from $\mathcal{D}_t$
   
      $s = s + 1$		 
		 
		 Compute the gradient $\overleftarrow g$ and $\overrightarrow g$
		 
		 \textit{\# preserve previous knowledge} 
		 
		 $\overleftarrow{W}^s_{1:(t-1)} = \overleftarrow{W}^{s-1}_{1:(t-1)} - \alpha \cdot P_{1:(t-1)} \cdot \overleftarrow g$ 
		 
		 \textit{\# learn new knowledge} 
	 
		 $\overrightarrow{W}^s_{t} = \overrightarrow{W}^{s-1}_{t} - \alpha \cdot \overrightarrow g$	 
	 }
	 \textit{\# linear connector}

	  $ W_{1:t} = \frac{t-1}{t} \overleftarrow{W}^s_{1:(t-1)} + \frac{1}{t} \overrightarrow{W}^s_{t}$
	  
	 \textit{\# compute the null space}
	
	Use the model $W_{1:t}$, $\mathcal{D}_t$ and $\mathcal{X}_{1:t-1}$ to obtain feature covariance $\mathcal{X}_{1:t}$ and the null-space projection matrix $P_{1:t}$	 
		}
		\KwOut{$W_{1:t}$}  
\end{algorithm}

\subsection{Linear Connector for Plasticity-Stability Trade-off}
According to the upper bound in Eq.~(\ref{upperBound2}),  we train two independent neural networks, which separately consider plasticity and stability, and the two models are moved towards each other. Finally, we design a simple linear connector according to linear connectivity and Eq.~(\ref{centroid}). 

\subsubsection{Remembering Knowledge of Previous Tasks} 

As discussed in \textbf{Section \ref{conlusion}}, the deep network considering stability should preserve the knowledge of past tasks and move towards the optimal set of the current task. 

Specially, we use Adam-NSCL to achieve the above goals. The previous model $W_{1:(t-1)}$ is used as the initialization of the deep network $\overleftarrow{W}_{1:(t-1)}$. 
 At iteration $s$, we randomly sample a mini-batch $\{X, Y\}$ from $\mathcal{D}_t$, and cross-entropy loss function is used to learn the model. The objective function can be formulated as
\begin{equation}
    \min_{\overleftarrow{W}_{1:(t-1)}} {\mathcal L}_{CE}(\overleftarrow{W}_{1:(t-1)}).
\end{equation}
We calculate the gradient as $\overleftarrow g$. To preserve the previous knowledge, the gradient is multiplied by the null-space projection matrix. The feature covariance of all $t-1$ tasks is $\mathcal{X}_{1:(t-1)}$ and the projection matrix of all previous data is $P_{1:(t-1)} = U U^{\top}$, where $U$ is the set of eigenvectors of $\mathcal{X}_{1:(t-1)}$ and their eigenvalues are zero. (Please refer to Algorithm 2 in~\cite{wang2021training} for more details of obtaining the feature covariance and projection matrix). Then, the weight is updated as
\begin{equation}
  \overleftarrow{W}^s_{1:(t-1)} = \overleftarrow{W}^{s-1}_{1:(t-1)} - \alpha \cdot P_{1:(t-1)} \cdot \overleftarrow g,
  \label{previous}
\end{equation}
where $\overleftarrow{W}^0_{1:(t-1)} = W_{1:(t-1)}$ and $\alpha$ is the stepsize. This updating strategy can ensure that it preserves the knowledge of past tasks, and also the model is moved towards the optimal set of the current task~\cite{wang2021training} .

\subsubsection{Learning New Knowledge of Current Task}
Here we update another deep network which considers plasticity. As discussed in \textbf{Section \ref{conlusion}}, the new model $\overrightarrow{W}_t$ should 1) be the optimal model of current task (the first term in the right of Eq.~(\ref{upperBound2})), and 2) be closer to the previous model (the second term in the right of Eq.~(\ref{upperBound2})). 

Specially, given the $t$-th training dataset $\mathcal{D}_t$, the objective function can be formulated as
\begin{align}
  \min_{\overrightarrow{W}_t} {\mathcal L}_{CE}(\overrightarrow{W}_t) + \mathcal{L}_{D}(\overrightarrow{W}_t), 
\label{fd}
\end{align}
where ${\mathcal L}_{CE}(\overrightarrow{W}_t)$ is the cross-entropy loss function aiming to learn the optimal weight of current task, and $\mathcal{L}_{D}(\overrightarrow{W}_t)$ aims to move the new model closer to the previous tasks. In general, the previous model $\overleftarrow{W}_{1:(t-1)}$ is not the optimal solution for the current task. Hence, simply using $\mathcal{L}_{D}(\overrightarrow{W}_t) = ||\overrightarrow{W}_{t} - \overleftarrow{W}_{1:(t-1)}||^2$ would hurt the performance of ${\mathcal L}_{CE}(\overrightarrow{W}_t)$. Instead, we use the feature distillation loss~\cite{zhang2020task}, which is formulated as
\begin{equation}
  \mathcal{L}_{D}(\overrightarrow{W}_t) = \frac{1}{|\mathcal{D}_t|}\sum_{\{X,Y\} \thicksim \mathcal{D}_t}\parallel F_{new}(X) - F_{old}(X) \parallel ^2,
\end{equation}
where $F_{new}$/$F_{old}$ are the feature extractors of $\overrightarrow{W}_t$/$W_{1:(t-1)}$, respectively. Please note that $\overrightarrow{W}_t$ consists of the feature extractor $F_{new}$ followed by the classifier $C_{new}$. And $F_{new}(X)$/$F_{old}(X)$ are the features of $X$ extracted by $F_{new}$/$F_{old}$, respectively. In this way, we can move the current model towards the previous tasks. 

Given the previous model $W_{1:(t-1)}$ as the initialization, we can simply use SGD or Adam~\cite{kingma2014adam} to learn the knowledge of current task and the gradient is $\overrightarrow g$. At iteration $s$, the neural network is updated as
\begin{equation}
  \overrightarrow{W}^s_{t} = \overrightarrow{W}^{s-1}_{t} - \alpha \cdot \overrightarrow g,
 \label{current}
\end{equation}
where $\overrightarrow{W}^0_{t} = W_{1:(t-1)}$. 

\subsubsection{Plasticity-Stability Trade-off}
Now we have two neural networks: $\overleftarrow{W}_{1:(t-1)}$ and $\overrightarrow{W}_{t}$. The $\overleftarrow{W}_{1:(t-1)}$ preserves the previous knowledge, and the $\overrightarrow{W}_{t}$ is the optimal weight of the current task. 
Formally, the linear connector between $\overleftarrow{W}_{1:(t-1)}$ and $\overrightarrow{W}_{t}$ is formulated as
\begin{equation}
(1-\beta) \overleftarrow{W}_{1:(t-1)} + \beta \overrightarrow{W}_{t},
\label{beta}
\end{equation} 
for $\beta \in [0,1]$. According to Eq.(~\ref{centroid}), we set $\beta = \frac{1}{t}$, which means that we average the weights of all $t$ tasks and get the final network as 
 \begin{equation}
  W_{1:t} = \frac{t-1}{t} \overleftarrow{W}_{1:(t-1)} + \frac{1}{t} \overrightarrow{W}_{t}.
\label{fusion}
\end{equation}
 The averaging model $W_{1:t}$ can achieve notable improvement. Stochastic weight averaging~\cite{izmailov2018averaging} also used the averaging model, and they showed that such averaging model can converge to the wider solution with better generalization. Our method is summarized in Algorithm ~\ref{alg}.

\textbf{Linear interpolation:} 
According to Observation~\ref{ob2}, to make the two networks linearly connected, we firstly use $W_{1:(t-1)}$ as the initialization to update two models. Then, the two networks are trained in similar manners to arrive the optima. 
The linear connector provides us a simple method to control the balance between the forgetting and intransigence by changing the value of $\beta : W_{1:t} = (1-\beta) \overleftarrow{W}_{1:(t-1)} + \beta \overrightarrow{W}_{t}$. If $\beta=0$, our method becomes Adam-NSCL, which mainly focuses on remembering knowledge of previous tasks. When $\beta=1$, it achieves excellent performance on the new task. Figure~\ref{ablation10} shows the performances of the linear combinations with different $\beta$.

\section{Experimental Results}
In this section, we evaluate our model on various incremental learning tasks and compared it with several state-of-the-art baselines. 
Besides, we have conducted ablation study to see the performances of previous and current tasks with various $\beta$ in \textbf{Eq. (\ref{beta})}. 
And we also evaluate our model using the evaluation measures of stability and plasticity.

\subsection{Datasets}

\textbf{CIFAR-100}~\cite{krizhevsky2009learning} is a dataset including 100 classes of images with size of $32 \times 32$ and each class contains 500 images for training and 100 images for testing.   \textbf{TinyImageNet}~\cite{wu2017tiny} contains 120,000 images of 200 classes. The images are downsized to $64 \times 64$ and each class contains 500 training images, 50 validation images and 50 test images. In this paper, the validation set of TinyImageNet is used for testing since the labels of test set are unavailable.

We split the dataset into $K$ disjoint subsets of classes such that the training samples of each task are from a disjoint subset of $C/K$ classes, where $C$ is the total number of classes and $K$ is the total number of tasks. When $K=10$, we get \textbf{10-split-CIFAR-100}, and the labels for 10 tasks are $\{\{0-9\},\{10-19\},..,\{90-99\}\}$, respectively. When $K=20$ and $K=25$, we get \textbf{20-split-CIFAR-100} and \textbf{25-split-TinyImageNet} respectively in the same way. In task $T_t$, we only have access to $\mathcal{D}_t$ and no previous data is stored. 

\subsection{Implementation Detail}

To make a fair comparison, we follow the experimental settings of Adam-NSCL~\cite{wang2021training}. Specifically, we use ResNet-18 as our backbone network and each task has its own single-layer linear classifier. When training in new task, we only update backbone network and classifier of new task, while classifiers of previous tasks remain unchanged. We use Adam optimizer, and the initial learning rate is set to $10^{-4}$ for the first task $T_1$ and $5 \times 10^{-5}$ for both $\overleftarrow{W}^0_{1:(t-1)}$ and $\overrightarrow{W}^0_{t}$ in other tasks. The total number of epochs is 80 and the learning rate is reduced by half at epoch 30 and epoch 60. The batch size for 20-split-CIFAR-100 is set to 32 and 16 for another two datasets.
For parameters that can not be updated by gradient descent method, e.g., $running\_mean$ of batch normalization layer, we also average them as \textbf{Eq. (\ref{fusion})}.

\subsection{Evaluation Protocol}
We use Average Accuracy (ACC) to measure how the model performs on all tasks. Here we denote the number of tasks as $K$. After finishing training from task $T_1$ to task $T_m$, the accuracy of model on test set of task $t$ is denoted as $A_{m,t}$. ACC can be calculated as
\begin{equation}
    {\rm ACC} = \frac{1}{K}\sum\limits_{t=1}^{K}A_{K,t},
\end{equation}
where $K$ is the total number of tasks. The larger ACC is, the better the model performs. Since it's the average accuracy of all tasks, we must take the balance between tasks into account.

We use Backward Transfer (BWT)~\cite{lopez-paz2017gradient} to measure how much the model forgets in the continual-learning process. BWT is defined as
\begin{equation}
    {\rm BWT} = \frac{1}{K-1}\sum\limits_{t=1}^{K-1}A_{K,t} - A_{t,t}.
\end{equation}
It indicates the average accuracy drop of all previous tasks. The larger BWT is, the less model forgets.  In this paper, we aim to achieve a more balanced model. 
Hence, ACC and BWT should be considered together. Given ACC and BWT two measures, we should firstly see the ACC: the larger value of ACC is better.  When the two methods have the same ACC values, we can use the BWT to observe how two methods perform on stability and plasticity: the smaller BWT means the method is good at learning new knowledge but it forgets more, larger BWT means it forgets less but learns less new task. 

\subsection{Results}
In this set of experiments, we compare our method with several state-of-the-art baselines. 
We compare our method with EWC~\cite{kirkpatrick2017overcoming}, MAS~\cite{aljundi2018memory} , MUC-MAS~\cite{liu2020more}, SI~\cite{zenke2017continual}, LwF~\cite{li2018learning}, InstAParam~\cite{chen2020mitigating}, GD-WILD~\cite{lee2019overcoming}, GEM~\cite{lopez-paz2017gradient}, A-GEM~\cite{chaudhry2018efficient}, MEGA~\cite{guo2020improved}, OWM~\cite{zeng2019continual} and Adam-NSCL~\cite{wang2021training}. All methods use ResNet-18 as backbone network for a fair comparison. 

\begin{table}[h]
\centering
\begin{tabular}{ccc}
\hline
Methods            & ACC(\%) & BWT(\%)  \\ \hline
EWC         & 70.77 & -2.83  \\
MAS         & 66.93 & -4.03  \\
MUC-MAS     & 63.73 & -3.38  \\
SI          & 60.57 & -5.17  \\
LwF         & 70.70 & -6.27  \\
InstAParam & 47.84 & -11.92 \\
GD-WILD    & 71.27 & -18.24 \\
GEM         & 49.48 & 2.77   \\
A-GEM       & 49.57 & -1.13  \\
MEGA       & 54.17 & -2.19  \\
OWM         & 68.89 & -1.88  \\
Adam-NSCL   & 73.77 & -1.6   \\ \hline
Ours & \textbf{79.79} & -0.92\\ \hline
\end{tabular}
\caption{Results on 10-split-CIFAR-100. Please note that a larger value of ACC is better. 
}
\label{res1}
\end{table}

\begin{table}[h]
\centering
\begin{tabular}{ccc}
\hline
Methods            & ACC(\%)          & BWT(\%)  \\ \hline
EWC         & 71.66          & -3.72  \\
MAS         & 63.84          & -6.29  \\
MUC-MAS     & 67.22          & -5.72  \\
SI          & 59.76          & -8.62  \\
LwF         & 74.38          & -9.11  \\
InstAParam & 51.04          & -4.92  \\
GD-WILD     & 77.16          & -14.85 \\
GEM        & 68.89          & -1.2   \\
A-GEM       & 61.91          & -6.88  \\
MEGA        & 64.98          & -5.13  \\
OWM         & 68.47          & -3.37  \\
Adam-NSCL   & 75.95          & -3.66  \\ \hline
Ours & \textbf{80.80} & -5.00 \\ \hline
\end{tabular}
\caption{Results on 20-split-CIFAR-100. A larger value of ACC is better and a moderate value of BWT is better for balanced model.}
\label{res2}
\end{table}

\begin{table}[h]
\centering
\begin{tabular}{ccc}
\hline
Methods        & ACC(\%)          & BWT(\%)  \\ \hline
EWC         & 52.33          & -6.17  \\
MAS         & 47.96          & -7.04  \\
MUC-MAS     & 41.18          & -4.03  \\
SI          & 45.27          & -4.45  \\
LwF         & 56.57          & -11.19 \\
InstAParam & 34.64          & -10.05 \\
GD-WILD     & 42.74          & -34.58 \\
A-GEM       & 53.32          & -7.68  \\
MEGA        & 57.12          & -5.90  \\
OWM         & 49.98          & -3.64  \\
Adam-NSCL   & 58.28          & -6.05  \\ \hline
Our & \textbf{64.61} &-6.00 \\\hline
\end{tabular}
\caption{Results on 25-split-TinyImageNet.}
\label{res3}
\end{table}

\begin{figure*}[ht]
    \centering
    \includegraphics[scale=1.2]{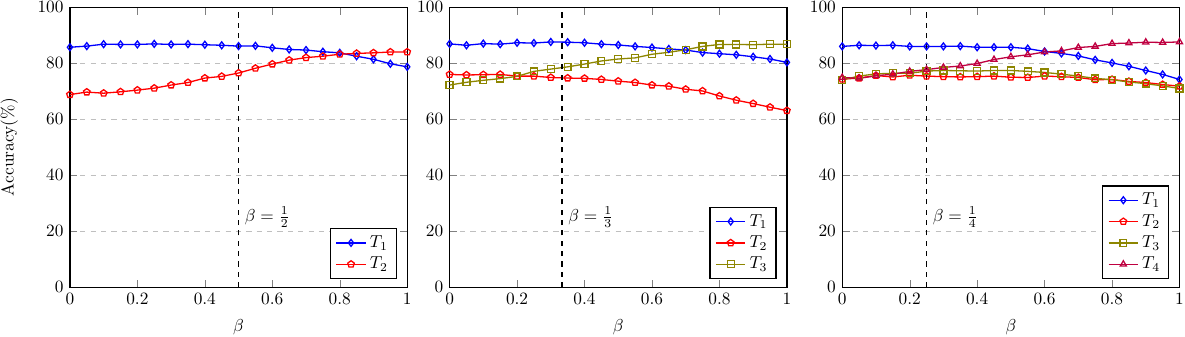}
    \caption{Accuracy of $W_{1:2}$(left), $W_{1:3}$(middle) and $W_{1:4}$(right) with different $\beta$ on 10-split-CIFAR-100}
    \label{ablation10}
\end{figure*}

\begin{figure*}[ht]
    \centering
    \includegraphics[scale=1.2]{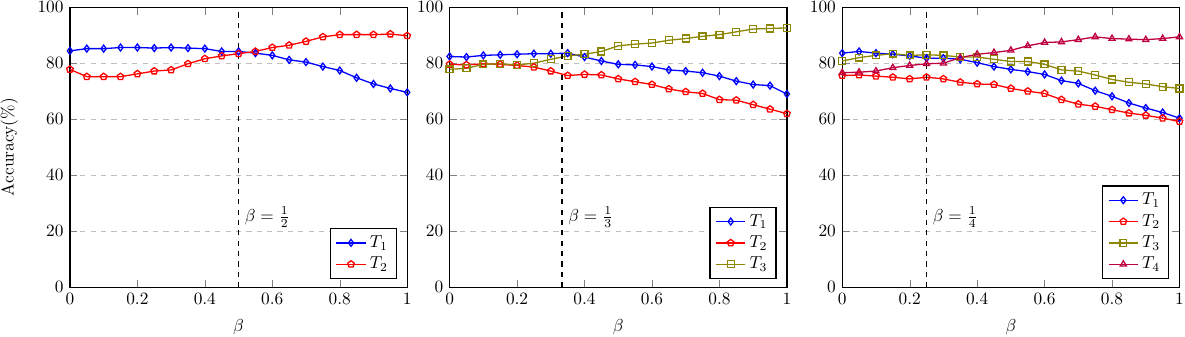}
    \caption{Accuracy of $W_{1:2}$(left), $W_{1:3}$(middle) and $W_{1:4}$(right) with different $\beta$ on 20-split-CIFAR-100}
    \label{ablation20}
\end{figure*}

\begin{figure*}[ht]
    \centering
    \includegraphics[scale=1.2]{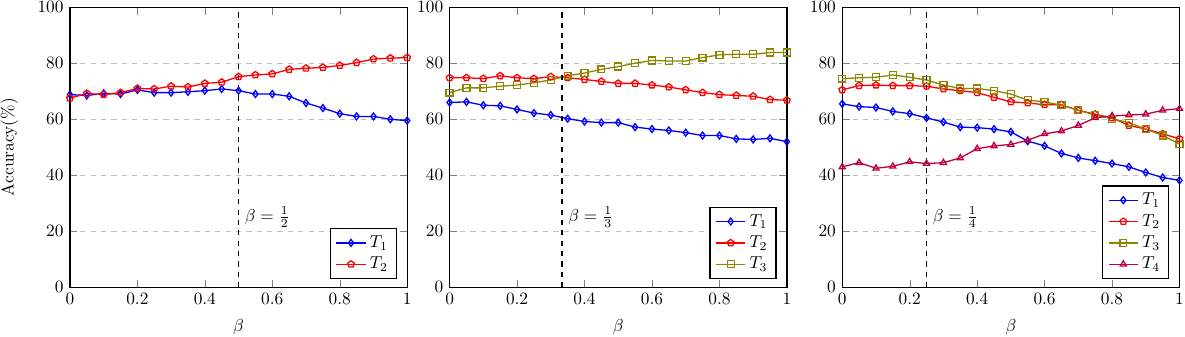}
    \caption{Accuracy of $W_{1:2}$(left), $W_{1:3}$(middle) and $W_{1:4}$(right) with different $\beta$ on 25-split-TinyImageNet}
    \label{ablation25}
\end{figure*}

Table~\ref{res1}, Table~\ref{res2} and Table~\ref{res3} show the comparison results. The results show that our method achieves significant improvement w.r.t. ACC on three datasets. The results of BWT and ACC indicate that our method can achieve better plasticity-stability trade-off. Detailed analysis is as follows.

\textbf{10-split-CIFAR-100}
The results are shown in Table~\ref{res1}. We can see that our model achieves the best ACC 79.79\%, which is 6.02\% superior to the second best model Adam-NSCL. The BWT value of our model is -0.92\%, which is the second best one compared to baselines. It indicates that our model can obtain a meaningful balance between previous tasks and new task.

\textbf{20-split-CIFAR-100} As shown in Table~\ref{res2}, our model still achieves the best ACC 80.80\%, which is 3.64\% better than the second best model GD-WILD. Note that GD-WILD stores previous data and its BWT value is 9.85\% worse than ours. Again, our model achieves a relative balanced BWT value -5.00\%. 

\textbf{25-split-TinyImageNet}
The results of Table~\ref{res3} show that our method achieves the best ACC 64.61\%, and the ACC of second best model Adam-NSCL is 58.28\%. The BWT and ACC  indicate that the our method not only can achieve better performance, but also obtain a more balanced model. Note that Adam-NSCL achieves an excellent performance, even so, our method performs better than Adam-NSCL. 

In summary, two observations can be made from the results: 1) our method yields the best performance on all datasets. 2) Our method achieves a better trade-off between stability and plasticity. Please note that the performance (ACC) of a IL model can be divided into two parts: stability (BWT) and plasticity. Hence, knowing ACC and BWT, we can probably know the performance of plasticity. We will further discuss it in the next subsection.

\subsection{Ablation Study}

In this set of experiments, we conduct ablation study on three benchmark datasets to see the effects of $\beta$. As indicated by \textbf{Eq. (\ref{beta})}, we use $\beta$ to control the ratio of two independent neural networks. 

For demonstration purposes, we only show three sequential learning tasks. The results of other tasks are similar. To be specific, when $t=2$, $T_1$ is the previous task and $T_2$ is the current task. The test accuracies of $W_{1:2}$ using different values of $\beta$ on tasks $T_1$ and $T_2$ are shown in the left of Figure~\ref{ablation10}, Figure~\ref{ablation20} and Figure~\ref{ablation25}. The results of $T_1$ indicate the ability to preserve old knowledge, and the accuracies of $T_2$ indicate the ability to learn new task. 

When $t=3$, test accuracies of tasks $T_1$, $T_2$ and $T_3$ are shown in the middle of Figure~\ref{ablation10}, Figure~\ref{ablation20} and Figure~\ref{ablation25}. 

When $t=4$, test accucaries of $T_1$, $T_2$, $T_3$ and $T_4$ are shown in the right of Figure~\ref{ablation10}, Figure~\ref{ablation20} and Figure~\ref{ablation25}. 

As shown in Figure~\ref{ablation10}, Figure~\ref{ablation20} and Figure~\ref{ablation25}, we can see that: 1) when $\beta = 0$, $W_{1:t} = \overleftarrow{W}_{1:(t-1)}$ can well preserve the previous knowledge. 2) When $\beta = 1$, $W_{1:t} = \overrightarrow{W}_{t}$ performs well on the new task. 3) The linear paths between $\overleftarrow{W}_{1:(t-1)}$ and $\overrightarrow{W}_{t}$ are almost smooth, and there are no obvious jumps along the paths. For example, the accuracy of $T_2$ increases as the value of $\beta$ gets larger as shown in the left of Figure~\ref{ablation10}. 4) For 10-split-CIFAR-100 and 20-split-CIFAR-100, the model strikes a balance well on all tasks when $\beta$ is close to $\frac{1}{t}$. For 25-split-TinyImageNet, though the fused model doesn't perform best on some tasks, $\beta = \frac{1}{t}$ is still the most compromising solution.

\subsection{Plasticity-Stability Trade-off Analysis}\label{analysis}
To better understand our method, we compare it with Adam-NSCL to analyse the plasticity-stability trade-off. 

We use BWT as the evaluation measure of stability. Further, we also use Intransigence Measure(IM)~\cite{chaudhry2018riemannian} to measure plasticity, which indicates how much the model has learnt from new task. The intransigence for the $k$-th task can be calculated as
\begin{equation}
    I_k = A_k^* - A_{k,k},
\end{equation}
where $A_k^*$ is the accuracy on the test set of $k$-th task with dataset $\cup_{i=1}^{k} \mathcal{D}_{i}$. The smaller the $I_k$ is, the better the model is.

Table~\ref{cmp1}, Table~\ref{cmp2} and Table~\ref{cmp3} show the results of BWT and IM. First, the BWT values of our model are bigger than that of Adam-NSCL except for 20-split-CIFAR-100, which means that Adam-NSCL has stronger ability to remember the previous knowledge for 20-split-CIFAR-100. Second, the IM values of our model are much better than Adam-NSCL.  Our method considers both the stability and plasticity, and the overall effect makes the ACC higher.  

\begin{table}[h]
\centering
\begin{tabular}{ll|ll}
\hline
Methods   & ACC & BWT(\%)       & $I_{10}$(\%)         \\ \hline
Adam-NSCL & 73.77 & -1.6 & 14.50         \\
Ours &\textbf{79.79} & \textbf{-0.92} & \textbf{8.10} \\ \hline
\end{tabular}
\caption{BWT and IM on 10-split-CIFAR-100}
\label{cmp1}
\end{table}

\begin{table}[h]
\centering
\begin{tabular}{ll|ll}
\hline
Methods   & ACC & BWT(\%)        & $I_{20}$(\%)         \\ \hline
Adam-NSCL & 75.95  & \textbf{-3.66} & 12.60         \\
Ours &\textbf{80.80} & -5.00 & \textbf{7.00} \\ \hline
\end{tabular}
\caption{BWT and IM on 20-split-CIFAR-100}
\label{cmp2}
\end{table}

\begin{table}[h]
\centering
\begin{tabular}{ll|ll}
\hline
Methods   & ACC & BWT(\%)        & $I_{25}$(\%)         \\ \hline
Adam-NSCL & 58.28 & -6.05 & 10.50         \\
Ours & \textbf{64.61} & \textbf{-6.00} &\textbf{5.75}\\ \hline
\end{tabular}
\caption{BWT and IM on 25-split-TinyImageNet}
\label{cmp3}
\end{table}

\section{Conclusion}
In this paper, we proposed a simple linear connector for incremental learning, which is a better plasticity-stability trade-off solution. To explain why we can use a simple linear connector to combine two models, we had given an analysis and showed it can minimize the upper bound of  empirical loss for all tasks. Hence, we proposed two independent neural networks. The first network aims to preserve the previous knowledge and the second network is to learn new knowledge. We used the null-space projection to learn the first network and the SGD for the second network. Finally, we simply averaged the two network and achieved a significant improvement. In our future work, we aim to find a better way to combine the two networks and give a better theoretical explanation for non-linear/linear connector.

\section*{Acknowledgment}
This work is supported by the National Natural Science Foundation of China under Grants (U1811261,U1811262), Guangdong Basic and Applied Basic Research Foundation (2019B1515130001, 2021A1515012172) and Zhuhai Industry-University-Research Cooperation Project (ZH22017001210010PWC).

{\small

\bibliographystyle{ieee_fullname}
}

\end{document}